%% file: main.tex
\documentclass[acmsmall]{acmart}

\AtBeginDocument{%
  \providecommand\BibTeX{{%
    \normalfont B\kern-0.5em{\scshape i\kern-0.25em b}\kern-0.8em\TeX}}}

\setcopyright{acmcopyright}
\copyrightyear{2021}
\acmYear{2021}
\acmDOI{xx.xxxx/xxxxxxx.xxxxxxx}

\acmJournal{JACM}
\acmVolume{}
\acmNumber{}
\acmArticle{}
\acmMonth{3}

\usepackage{booktabs} 
\usepackage{soul}
\usepackage{url} 
\usepackage{graphicx} 
\usepackage{latexsym}
\usepackage{CJK}
\usepackage{amsmath}
\usepackage{bm}
\usepackage{mathrsfs}
\usepackage{float}
\usepackage{multirow}
\usepackage{xcolor,colortbl} 
\usepackage{subfigure}
\usepackage{bigstrut,bigdelim}
\usepackage{algorithm}  
\usepackage{algpseudocode}  
\usepackage{lipsum}
\usepackage{bbding}
\usepackage{pifont}

\begin{document}

\title{Dialogue History Matters! Personalized Response Selection in Multi-turn Retrieval-based Chatbots}


\author{Juntao Li}
\authornote{Equal contribution. Ordering is decided by a coin flip.}
\email{lijuntao@pku.edu.cn}
\author{Chang Liu}
\authornotemark[1]
\email{liuchang97@pku.edu.cn}
\affiliation{%
  \institution{Wangxuan Institute of Computer Technology and Center for Data Science, Academy for
Advanced Interdisciplinary Studies, Peking University}
  \streetaddress{\#5 Yiheyuan Rd}
  \city{Haidian Qu}
  \state{Beijing Shi}
  \country{China}}

\author{Chongyang Tao}
\affiliation{%
  \institution{Wangxuan Institute of Computer Technology, Peking University}
  \streetaddress{\#5 Yiheyuan Rd}
  \city{Haidian Qu}
  \state{Beijing Shi}
  \country{China}}
\email{chongyangtao@pku.edu.cn}

\author{Zhangming Chan}
\affiliation{%
  \institution{Wangxuan Institute of Computer Technology, Peking University}
  \streetaddress{\#5 Yiheyuan Rd}
  \city{Haidian Qu}
  \state{Beijing Shi}
  \country{China}}
\email{zhangming.chan@pku.edu.cn}

\author{Dongyan Zhao}
\authornote{Corresponding author.}
\affiliation{%
  \institution{Wangxuan Institute of Computer Technology, Peking University}
  \streetaddress{\#5 Yiheyuan Rd}
  \city{Haidian Qu}
  \state{Beijing Shi}
  \country{China}
}
\email{zhaody@pku.edu.cn}

\author{Min Zhang}
\affiliation{%
  \institution{Soochow University}
  \streetaddress{\#1 Shizi Rd}
  \city{Suzhou}
  \state{Jiang Su}
  \country{China}
}
\email{minzhang@suda.edu.cn}

\author{Rui Yan}
\affiliation{%
 \institution{Gaoling School of Artificial Intelligence, Renmin University of China}
  \streetaddress{\#59 Zhongguancun Rd}
  \city{Haidian Qu}
  \state{Beijing Shi}
  \country{China}}
\email{ruiyan@ruc.edu.cn}

\renewcommand{\shortauthors}{Juntao Li and Chang Liu, et al.}

\begin{abstract}
Existing multi-turn context-response matching methods mainly concentrate on obtaining multi-level and multi-dimension representations and better interactions between context utterances and response.
However, in real-place conversation scenarios, whether a response candidate is suitable not only counts on the given dialogue context but also other backgrounds, e.g., wording habits, user-specific dialogue history content.
To fill the gap between these up-to-date methods and the real-world applications, we incorporate user-specific dialogue history into the response selection and propose a personalized hybrid matching network (PHMN).
Our contributions are two-fold: 
1) our model extracts personalized wording behaviors from user-specific dialogue history as extra matching information; 
2) we perform hybrid representation learning on context-response utterances and explicitly incorporate a customized attention mechanism to extract vital information from context-response interactions so as to improve the accuracy of matching.
We evaluate our model on two large datasets with user identification, i.e., personalized Ubuntu dialogue Corpus (P-Ubuntu) and personalized Weibo dataset (P-Weibo).
Experimental results confirm that our method significantly outperforms several strong models by combining personalized attention, wording behaviors, and hybrid representation learning.
\end{abstract}



\maketitle

\input{1-introduction}
\input{2-preliminary}
\input{3-model}
\input{4-experiments}
\input{5-results}
\input{6-related_work}

\section{Conclusion}
In this study, we propose a novel personalized hybrid matching network (PHMN) for multi-turn response selection through leveraging user-specific dialogue history as extra information.
Building upon the advanced multi-dimension hybrid representation learning strategy, we incorporate the information in dialogue history from various granularities, i.e., wording behaviors matching, user-level attention for extracting vital matching information from context-response matching.
Experimental results on two large datasets with different languages, personalized Ubuntu dialogue corpus (P-Ubuntu), and personalized Weibo (P-Weibo), confirm that our proposed method significantly outperforms state-of-the-art models (without using BERT).
We also conduct a thorough ablation study to investigate the effect of wording behavior modeling and the influence of personalized attention, which confirms that both wording behavior and personalized attention are effective for enhancing context-response matching.
Besides, we further explored the influence of Speaker A's persona in conversation insomuch as individuals will perform distinctive behaviors when they have a chat with different people.
In the near future, we pursue to learn a structured knowledge representation of users and encapsulate this structured information into response selection.

\begin{acks}
We would like to thank the efforts of anonymous reviewers for improving this paper. 
This work was supported by the National Key Research and Development Program of China (No.2020AAA0106600).
\end{acks}


\bibliographystyle{ACM-Reference-Format}


\end{document}

%% file: 1-introduction.tex
\section{Introduction}
Dialogue systems have received a considerable amount of attention from academic researchers and have achieved remarkable success in a myriad of industry scenarios, such as in chit-chat machines \cite{shum2018eliza}, information seeking and searching \cite{aliannejadi2019asking,Hashemi2020GuidedTL}, and intelligent assistants \cite{li2017alime}.
From the perspective of domains involved in previous studies, existing studies can be categorized into two groups, i.e., domain-specific and open-domain.
Domain-specific models generally pursue to solve and complete one specific target (e.g., restaurant reservation \cite{lei2018sequicity}, train routing \cite{ferguson1996trains}), which always involve domain knowledge and engineering.
Unlike domain-specific studies, open-domain dialogues between human and machine involve unlimited topics within a conversation \cite{ritter2011data}, as a result of which building an open-domain dialogue system is more challenging with the lack of enough knowledge engineering.
Benefiting from the explosion of available dialogue datasets, constructing open-domain dialogue systems has attracted a growing number of researchers.
Among dialogue systems in open-domain, generation-based \cite{shangL2015neural,sordoni2015neural,vinyals2015neural} and retrieval-based \cite{wang2013dataset,wu2018learning,wu2017sequential} methods are the mainstreams in both academia and industry, where generation methods learn to create a feasible response for a user-issued query while retrieval-based methods extract a proper response from a candidate pool.
In contrast to the ``common response''\footnote{Sequence-to-sequence neural networks along with the log-likelihood objective function tend to create short, high-frequency, and commonplace responses (e.g., ``I don't know'', ``I'm OK''), which also refers to common response in previous study \cite{sordoni2015neural,vinyals2015neural,serban2016building}} created by generation models \cite{li2015diversity}, retrieval-based methods can extract fluent and informative responses from human conversations \cite{tao2019multi}.
Early retrieval-based methods mainly address the issue of single-turn response selection, where the dialogue context only contains one utterance \cite{wang2015syntax}.
Recent studies focus on modeling multi-turn response selection \cite{lowe2015ubuntu}.

For multi-turn response selection, a dialogue system is required to properly calibrate the matching degree between a multi-turn dialogue context and a given response candidate.
The response selection task thus can be naturally transformed to learning the matching degrees between the semantic representations and the dependency relationships between context and response candidates.
SMN (sequential matching network) attempts to learn fine-grained (e.g., word-level, sentence-level) semantic matching information between each utterance in context and the response candidate and aggregate the matching information to calculate the final matching results.
MIX (multi-channel information crossing) \cite{chen2018mix} models the matching degrees between context and response from the perspective of interaction representations \cite{tao2019multi} to extract multi-channel matching patterns and information.
Deep attention matching network (DAM) \cite{zhou2018multi} captures sophisticated dependency information in utterances and cross utterances, i.e., using self-attention mechanism and cross-attention strategy to learn the representations of context and response candidate.
Although these methods have achieved promising results, there is still room for improving their capability of context utterance modeling, such as combining dependency relationships and multi-channel interaction representations \cite{chen2018mix,tao2019multi}. 
Besides, existing models are trapped into learning matching signals from context and response, leaving introducing extra information unexplored.
Table \ref{tab:intro_case} illustrates a sampled case in our experiments.
We can observe that the word ``\textit{xfce4}'' is the crucial clue for selecting the target response.
However, other words are likely to overwhelm the word ``\textit{xfce4}'', leading to unsatisfactory performance as it appears only once in the context.
If we exploit the information in history-response matching, i.e., the high-frequency word ``\textit{xfce4}'', the performance of response selection will be enhanced.
A recent study \cite{yang2018response} proposes to use pseudo-relevant feedback documents as an extra information source to enrich the response representation.
Such a strategy is useful but still risky since the pseudo-relevant feedback might introduce much-unrelated information.
Thus, it is imperative to use more accurate extra information, i.e., user-specific dialogue history, for improving the matching performance.
\input{table-1}
To address these obstacles, we propose a novel personalized hybrid matching network (PHMN) for multi-turn response selection, in which the customized dialogue history is introduced as additional information, and multiple types of representations in context and response are merged as the hybrid representation.
Explicitly, we incorporate two types of representations in the hybrid representation learning, i.e., attention-based representation for learning dependency information, interaction-based representation for extracting multiple matching patterns and features.
Such a strategy has been proved to be effective in recent studies for improving context-response matching \cite{tao2019multi,tao2019one}.
In exploiting information in dialogue history, we introduce two different ways for enhancing multi-turn context response matching.
For one thing, we extract the wording behavior of a specific user in the corresponding dialogue history as long-term information to supplement the short-term context-response matching.
For another, we compute personalized attention weights for candidate responses to extract critical information for response selection.
More concretely, we perform the personalized attention upon the learned hybrid representation and then utilize a gate mechanism to fuse the wording behavior matching information and weighted hybrid context-response matching information.
Hence, our model is effective for both context modeling and dialogue history exploiting.

We conduct experiments on two challenging datasets for personalized multi-turn response retrieval, i.e., personalized Ubuntu dialogue corpus (P-Ubuntu) and personalized Weibo dataset (P-Weibo), to evaluate the effectiveness of our proposed PHMN model, where there exist user ids in both datasets.
Experimental results confirm that our model achieves state-of-the-art performance on the newly created corpora. 
Through introducing personalized wording behaviors and personalized attention, our model yields a significant improvement over several strong baseline models, which suggests that introducing user-specific dialogue history and learning hybrid representations are appealing for multi-turn response retrieval. 

%% file: table-1.tex
\begin{table}[t]
\caption{\label{tab:intro_case} An example from the raw Ubuntu dataset that illustrates dialogue history can benefit response matching. 
}
\begin{center}
\begin{tabular}{l|l}
\hline
 & B: i've read somewhere that \textcolor{blue}{\textbf{\textit{xfce4}}} is as fast as fluxbox \\
{\bf Dialogue } & B: i use \textcolor{blue}{\textbf{\textit{xfce4}}} , old laptop gnome runs terribly slow on it\\
{\bf History } & B: haven't tried kde on this laptop, but when i tried \textcolor{blue}{\textbf{\textit{xfce4}}} its like a new lease of life \\
& B:  \textcolor{blue}{\textbf{\textit{xfce4}}} is light, yet quite functional \\
\hline
\multirow{7}*{\bf Context} & A: do anyone know how to add shortcuts to the menu ? \\
  & B: depends on your desktop environment  \\
   & A: sorry i new in ubuntu, what do you mean with desktop enviroment?  \\
    & B: KDE / GNOME / \textcolor{red}{\textbf{\textit{xfce4}}}/ fluxbox ??  \\ 
    & A: its GNOME  \\
     & B: old laptop GNOME runs terribly slow on it  \\
      & A: umm yup.. what do you suggest then?   \\
\hline
\bf Target &B: Try \textcolor{red}{\textbf{\textit{xfce4}}} it's wonderfull, as light as icewm, and more confortable to use \\
\hline
\end{tabular}
\end{center}
\end{table}

%% file: 2-preliminary.tex
\section{Preliminaries}
\subsection{Deep Matching Network}
Generally, the recent effective deep matching networks for multi-turn response retrieval or short text matching consist of four elements: representations learning, dependency modeling, matching, aggregation and fusion. 

\textbf{\textit{Representation Learning.}}
Most studies for multi-turn response selection first transform context utterances and response candidates to either vector representations \cite{huang2013learning} or interaction matrices \cite{hu2014convolutional,pang2016text,guo2016deep} for convenient matching calculation.
For vector representations learning, various deep neural networks are designed for learning multi-level and multi-dimension semantic information from conversation utterances, including CNN-based \cite{kalchbrenner2014convolutional,kim2014convolutional}, RNN-based \cite{li2015tree,liu2016recurrent}, and tree-RNN-based methods \cite{irsoy2014deep,socher2011parsing}.
As to interaction-based representation learning methods, they first generate an interaction matrix for each utterance pair between context utterances and response candidates.
Then, direct matching features such as the degree and structure of matching are captured by a deep neural network \cite{xiong2017end,dai2018convolutional}.

\textbf{\textit{Dependency Modeling.}}
Besides the semantic representations and matching structures in the interaction-based method, there exist sophisticated dependency information and reference relations within utterances and across utterances. 
Benefited from the great success of the Transformer on neural machine translation, various attention-based methods are proposed to capture the dependency structure and information from different levels.
DAM \cite{zhou2018multi} leverages a novel attention-based multi-turn response selection framework for learning various dependency information and achieves very competitive performance.
DAM borrows the self-attention strategy from Transformer for capturing word-level intra-utterance dependency and sentence-level representations and uses a cross-attention mechanism to capture dependency (e.g., reference relations) between those latently matched segment pairs.

\textbf{\textit{Matching.}}
Once obtaining utterance representations at each level of granularity, the matching relations between two segments will be calculated.
According to the information in utterance representations, semantic matching methods and structure matching approaches are designed to calibrate the matching degree between two representations.
To date, various matching degree calculation methods have been investigated, e.g., using Euclidean distance between two vectors as the matching degree, performing cosine similarity calculation, computing element-wise dot production.
Based on the information in vector representations (semantic or structure) and different matching degree calibration strategies, an effective matching method can be designed for comprehensively computing the matching degree between two utterances.

\textbf{\textit{Aggregation and Fusion.}}
After calculating the matching degree between context and response at each level of granularity, a typical deep matching network contains an aggregation or fusion module for learning the final matching score.
SMN \cite{wu2017sequential} proposes to use RNN to sequentially accumulate the matching degree of each utterance-response pair and further compute the matching score between the context and the response.
As utterances relationships within a dialogue context have effects on the calculation of the final matching score, DUA \cite{zhang2018modeling} refines the utterance representations with gated self-attention and further aggregates this information into a matching score.
DAM \cite{zhou2018multi} aggregates all the matching degrees of segments across each utterance and response into a 3D tensor and then leverages two-layered 3D convolutions with max-pooling operations to fuse the matching degree information and compute the final matching score.

\subsection{Problem Formulation}
We follow the conventional settings in previous multi-turn response retrieval works \cite{tao2019one,tao2019multi} and introduce the following necessary notations to formulate the personalized multi-turn response retrieval task.
A dataset with user dialogue history content 
\begin{math}
\mathcal{D}=\{(c_i, r_i, m_i, y_i)\}_{i=1}^N
\end{math} 
is first given, where $c_i$, $r_i$, $m_i$, $y_i$ represent dialogue context, response candidate, dialogue history and the corresponding binary label of the response candidate respectively.
Note that we treat user dialogue history utterances as the extra information for building a personalized multi-turn dialogue response retrieval model.
For the sake of clarity, we omit the subscript $i$ which denotes the case index in $\mathcal{D}$ when elaborating the details of our model.
Herein, an utterance $c$ in the dialogue context is represented as 
\begin{math}
c=(u_1, u_2,\ldots, u_j,\ldots, u_{n_c})
\end{math} 
where $u_{j}$ represents an utterance with length $n_{u_{j}}$ in the j-th turn of the dialogue context and there are $n_c$ utterances in the dialogue context.
Similarly, there are $n_m$ history utterances of the current user who is supposed to raise a response for the given dialogue context, which is denoted as 
\begin{math}
m=(u_{m,1}, u_{m,2},\ldots, u_{m,k},\ldots, u_{m,n_m})
\end{math}, 
where $u_{m,k}$ represents an utterance with length $n_{u_{m,k}}$.
$n_r$ denotes the number of words in a candidate response $r$.
$y=1$ means the given response candidate is proper for the context and corresponded user dialogue history, otherwise $y=0$.
Then, our task is defined as learning a matching function $f(\cdot)$ from the given dataset that can yield a matching score between the dialogue context and the given response candidate with the help of user dialogue history.

%% file: 3-model.tex
\section{Model}
Inspired by the advanced deep multi-turn dialogue response selection framework mentioned above, we design our model from two directions, i.e., obtaining more comprehensive information from context and response, introducing more auxiliary information other than context and response.
We proposed a personalized hybrid matching network (PHMN) for multi-turn response selection, which incorporates hybrid representations in context and response (i.e., semantic matching, interaction-based features, and dependency relations information) and personalized dialogue content (i.e.,  user-specific dialogue history).
As shown in figure \ref{fig:framework}, our proposed PHMN comprises three main sub-modules, i.e., hybrid representation learning module, personalized dialogue content modeling, aggregation and fusion.

\subsection{Hybrid Representations Learning}
We consider obtaining semantic representations of context and response at two different levels, i.e., word-level and phrase-level.
Concretely, we adopt word embeddings as the word-level representations and the combination of uni-gram, bi-gram, tri-gram semantic information as phrase representations.
We also borrow the strategy of self-attention from the Transformer \cite{vaswani2017attention} and DAM \cite{zhou2018multi} to learn abundant dependency relationships in conversations. 
To capture the matching structure and patterns, we transform the semantic representations of context and response to interaction matrices.
Details of learning \textit{word representation}, \textit{phrase representation}, \textit{dependency representation} and constructing \textit{interaction matrices} are elaborated as follows:

\textbf{Word Representations.}
We use word embeddings as word-level representations insomuch as they contain rich semantic information and co-occurrence information. In learning, we initialized word embeddings with pre-trained Word2Vec on each benchmark dataset, i.e., P-Ubuntu dialogue corpus in English and P-Weibo dataset in Chinese. 
Upon both datasets, the dimension of word embedding is $d_w$.
Note that any proper Word2vec learning algorithm and pre-trained results are applicable such as BERT \cite{devlin2018bert}.
Thus, the word-level representation of an utterance $u_j$ is  
\begin{math}
{\bm{U_j}}=[\bm{e_{u_j, 1}}, \bm{e_{u_j, 2}}, \dots, \bm{e_{u_j, k}}, \dots, \bm{e_{u_j, n_{u_j}}}] \in \mathbb{R}^{n_{u_j}\times d_w}
\end{math}; and similarly a response candidate $r$ can be written as
\begin{math}
{\bm R}=[\bm{e_{r,1}}, \bm{e_{r,2}}, \dots, \bm{e_{r,k}}, \dots, \bm{e_{r,n_r}}]  \in \mathbb{R}^{n_{r}\times d_w}
\end{math}.
The dimensions of $\bm{e_{u_j, k}}$ and $\bm{e_{r,k}}$ are both $d_w$.

\input{figure-1}

\textbf{Phrase Representations.} 
In an actual situation, obtaining semantic representations solely based on word representations is risky as the semantic assemble patterns of words differ from each other.
For instance, ``all in'' and ``in all'' have totally different semantic information, while ``work hard'' and ``hard work'' deliver the same semantic content.
We consider modeling the semantic assemble patterns with a convolutional neural network.
In both English and Chinese, the minimal semantic unit typically includes 1 to 3 words \cite{chen2018mix}.
As a result, we conduct convolution operations upon the word embedding representations with different window sizes to capture uni-gram, bi-gram, and tri-gram information.
Concretely, we conduct 1-D convolution on the word embeddings of a given utterance \begin{math}
{\bm{U_j}}=[\bm{e_{u_j, 1}}, \bm{e_{u_j, 2}}, \dots, \bm{e_{u_j, k}}, \dots, \bm{e_{u_j, n_{u_j}}}]
\end{math} 
with window size $l$ from 1 to 3, where there are $d_f$ filters for each window size and the stride length is 1.
The $l$-gram phrase representation in the $k$-th location is calculated as:
\begin{equation}
\bm{{o_k}^l} = ReLU(\bm{{Z_k}^l}\bm{W_l} + \bm{b_l})
\end{equation}
where $\bm{W_l}$ and $\bm{b_l}$ are trainable parameters of the convolutional filter with window size $l$, and $\bm{{Z_k}^l} \in \mathbb{R}^{l \times d_w}$ stands for the input unigram embeddings in the current sliding window which is formulated as:
\begin{equation}
\bm{{Z_k}^l} = [\bm{{e_{k-\lfloor \frac{1}{2}(l-1) \rfloor}}}, \ldots,  \bm{{e_{k}}}, \ldots, \bm{{e_{k+\lfloor \frac{1}{2}l \rfloor}}}]
\end{equation}
where $\bm{{e_{k}}}$ is the word embedding representation of a word in either the dialogue context or the response (i.e., it can be either $\bm{e_{u_j, k}}$ or $\bm{e_{r, k}}$). Here we set $d_f$ the same as $d_w$. The output sequence of vectors of the convolution has the same length as the input sequence of vectors by utilizing the zero-padding strategy.
Thus, a given utterance $u_j$ is transformed to three matrix, i.e.,  ${\bm {U_j}^1}=[\bm{o^1_1}, \bm{o^1_2},\dots, \bm{o^1_{n_{u_j}}}]$, ${\bm {U_j}^2} =[\bm{o^2_1}, \bm{o^2_2},\dots, \bm{o^2_{n_{u_j}}}]$, ${\bm {U_j}^3}=[\bm{o^3_1}, \bm{o^3_2},\dots, \bm{o^3_{n_{u_j}}}]$. $\bm {U_j}^1$, $\bm {U_j}^2$ and $\bm {U_j}^3$ correspond to \{1,2,3\}-gram semantic information, respectively.
Similarly, we also conduct 1-D convolution on a given response \begin{math}
{\bm R}=[\bm{e_{r,1}}, \bm{e_{r,2}}, \dots, \bm{e_{r,k}}, \dots, \bm{e_{r,n_r}}]
\end{math} using the same convolutional filters, which outputs three matrix $\bm{R^1}$, $\bm{R^2}$, $\bm{R^3}$.

\textbf{Dependency Representations.}
To obtain the sophisticated dependency representations in conversations, we utilized an attentive module that is similar to the attention module in Transformer \cite{vaswani2017attention} and DAM.
The attentive module takes three sentences as input, namely the query sentence, the key sentence, and the value sentence, which are denoted as ${\bf \mathcal{Q}}=[\bm{e_i}]_{i=0}^{n_\mathcal{Q}-1}$, ${\bf \mathcal{K}}=[\bm{e_i}]_{i=0}^{n_\mathcal{K}-1}$, ${\bf \mathcal{V}}=[\bm{e_i}]_{i=0}^{n_\mathcal{V}-1}$ respectively, where $n_\mathcal{Q}$, $n_\mathcal{K}$, $n_\mathcal{V}$ represent the number of words in each sentence and $n_\mathcal{K}=n_\mathcal{V}$, and $\bm{e_i}$ is the $d_w$-dimension word embedding representation of a word.
The attentive module first uses each word in the query sentence to attend each word in the key sentence through the scaled dot-product attention mechanism.
Then, the obtained attention score is functioned upon the value sentence $\bf \mathcal{V}$ to form a new representation of $\bf \mathcal{Q}$, which is formulated as follows:
\begin{equation}
Att({\bf \mathcal{Q,K,V}})= softmax(\frac{{\bf \mathcal{QK}}^T}{\sqrt{d_w}})\bf \mathcal{V}
\end{equation}
In practice, the key sentence and the value sentence are identical, i.e., $\mathcal{K}=\mathcal{V}$.
Thus, each word in the query sentence $\mathcal{Q}$ is represented by the joint meaning of its similar words in $\mathcal{V}$.
We dispense $h$ heads to $\mathcal{Q}$, $\mathcal{K}$, $\mathcal{V}$ to capture multiple aspects dependency information via the scaled dot-product multi-head attention.
The output of head $i$ is then written by
\begin{equation}
{\bf \mathcal{O}_i}=Att({\mathcal{Q} \bm{W_i}^{\mathcal{Q}},{\mathcal{K}} \bm{W_i}^{\mathcal{K}},{\mathcal{V}} \bm{W_i}^{\mathcal{V}}})
\end{equation}
where $\bm{W_i}^\mathcal{Q}$, $\bm{W_i}^\mathcal{K}$, $\bm{W_i}^\mathcal{V}$ $\in \mathbb{R}^{d_w\times(d_w/h)}$ are trainable parameters for linear transformations.
The outputs of each head are concatenated to obtain the attention representations, formulated as:
\begin{equation}
\mathcal{O}=(\mathcal{O}_1\oplus \mathcal{O}_2\oplus \dots \oplus \mathcal{O}_h) \bm{W}_\mathcal{O}
\end{equation}
where $\oplus$ represents column-wise concatenation operation and $\bm{W_\mathcal{O}} \in \mathbb{R}^{d_w\times d_w}$ is trainable.
We then applied a layer normalization operation for preventing the vanishing or exploding of gradients.
We also use a residual connection to add the output $\mathcal{O}$ to the query sentence $\mathcal{Q}$.
From here, we denote the whole attentive module as $Attention({\bf \mathcal{Q,K,V}})$.
Note that the output of the attentive module has an identical dimension with the query sentence $\mathcal{Q}$.
In experiments, $\mathcal{Q,K,V}$ are set to same, i.e., $\mathcal{Q=K=V}$.
For a given context utterance $u_j$, its attention-based representation $\bm {U_j}^a$ is calculated as the output of $Attention(\bm{U_j},\bm{U_j},\bm{U_j})$.
In this way, an utterance can attend itself to represent each word with other related words within the utterance.
As a result, dependency relation information among the utterance can be captured.
Similarly, the dependency representation of a given response is ${\bm{R^a}}=Attention(\bm{R},\bm{R},\bm{R})$.

\textbf{Interaction Matrices.}
Given an utterance $u_j$ in a context and a response $r$, we have five-channel representations for $u_j$ and $r$ respectively, i.e., ${\bm{U_j},  \bm{{U_j}^1}, \bm{{U_j}^2}, \bm{{U_j}^3}, \bm{{U_j}^a}}$ and ${\bm{R}, \bm{R^1}, \bm{R^2}, \bm{R^3}, \bm{R^a}}$, where each representation channel of $u_j$ has a dimension of $\mathbb{R}^{n_{u_j}\times d_w}$ and each representation channel of $r$ has a dimension of $\mathbb{R}^{n_r\times d_w}$.
We then construct five interaction matrices for each utterance-response pair, which correspond to the interactions of $\bm{U_j}$-$\bm R$, $\bm {U_j}^1$-$\bm{R^1}$, $\bm {U_j}^2$-$\bm {R^2}$, $\bm {U_j}^3$-$\bm {R^3}$, $\bm {U_j}^a$-$\bm {R^a}$.
Take the interaction of $\bm U_j$-$\bm R$ as an example, the $(p,q)$-th element of the interaction matrix $\bm{M_j}$ is calculated by the dot production of the $p$-th element of $\bm R$ and the $q$-th element of $\bm {U_j}$. In practice, we directly use matrix multiplications to calculate each of the five interaction matrices. The calculation of $\bm {M_j}$ is as follows:
\begin{equation}
\bm{M_j}=\bm{R} \cdot \bm{U_j}^T
\end{equation}
Following the same calculation procedure, we can obtain the other four interaction matrices $\bm {M_j}^1, \bm{M_j}^2, \bm{M_j}^3, \bm{M_j}^a$, where each matrix has a dimension of $\mathbb{R}^{n_r \times n_{u_j}}$.

\subsection{Personalized Dialogue Content Modeling}
In addition to the hybrid representation of context and response, we propose to use the user-specific dialogue content from two perspectives.
For one thing, for a given dialogue context, the different user has distinctive attention to the context when matching a response candidate. In other words, some words or phrases are more important than others for response selection, and those vital content changes for different users. 
For another, we assume the user wording behavior in dialogue history is effective supplementary information for response selection.
Thus, personalized dialogue content modeling depends on how to learn personalized attention to allocate weight to each word or phrase in matching context utterances and response candidate, and how to extract wording behavior matching information between dialogue history and response candidate.

\textbf{Personalized Attention.}
\input{figure-2}
Intuitively, the words and phrases in context utterances and response candidate are not equally important for response selection. Moreover, different users may have distinctive attention to the dialogue content. To model the relative importance among the words and phrases with the consideration of the users' persona, we propose a simple but effective method for calculating the personalized attention scores from history utterances, which takes phrases distributions at multiple granularities into account.
We first construct the personalized TF-IDF corpus by treating the dialogue history of each user as a document. Then we can compute the $\{1,2,3\}$-gram TF-IDF scores for each given utterance. In doing so, each $\{1,2,3\}$-gram phrase in the response candidate is allocated with a weight. We then conduct these weights on the interaction matrices of each context utterance and response pair. 
Recall that we have $\bm{M_j}, \bm{{M_j}^1}, \bm{{M_j}^2}, \bm{{M_j}^3}, \bm{{M_j}^a}$ for each context utterance and response pair, representing interactions at the word embedding level, the uni-gram level, the bi-gram level, the tri-gram level, and the self-attention dependency.
Specifically, for the given response $r$, we calculate its $\{1,2,3\}$-gram personalized weights as $\bm{{a}^1}$, $\bm{{a}^2}$ and $\bm{{a}^3}$ whose dimensions are all $\mathbb{R}^{n_{r}\times 1}$. We then copy these score vectors $n_{u_j}$ times in the column direction to form the personalized mask matrices $\bm{{A}^1}$,  $\bm{{A}^2}$ and $\bm{{A}^3}$. All the three personalized mask matrices have the same dimension of $\mathbb{R}^{n_{r} \times n_{u_j}}$, and the values in the same row within a matrix are the same. As the rows of the interaction matrices represent the response, we directly multiply the $\{1,2,3\}$-gram personalized mask matrices to the corresponding $\{1,2,3\}$-gram interaction matrices. Concretely, we multiply $\bm {A}^1$ to $\bm {M_j}, \bm{{M_j}^1}, \bm{{M_j}^a}$, multiply $\bm {A}^2$ to $\bm {M_j}^2$, and multiply $\bm {A}^3$ to $\bm {M_j}^3$.
As shown in Figure \ref{fig:mask}, we denote these weights as the personalized masks to extract vital matching signals in the interaction matrices, resulting five new interaction matrices $\bm{{M_j}^{'}}, \bm{{M_j}^{1'}}, \bm{{M_j}^{2'}}, \bm{{M_j}^{3'}}, \bm{{M_j}^{a'}}$ for each context utterance response pair.

\textbf{Wording Behavior Matching.}
In analogy to the phrase representations of context utterances and response, we treat the $\{1,2,3,4\}$-grams matching information and patterns as wording behavior matching information.
In details, we conduct 1-D convolution on a response candidate ${\bm R}=[\bm{e_{r,1}}, \bm{e_{r,2}}, \ldots, \bm{e_{r,n_r}}]$, and a history utterance ${\bm{U_{m,k}}}=[\bm{e_{{m,k}, 1}}, \bm{e_{{m,k}, 2}},$ \ldots, $\bm{e_{m, n_{u_{m,k}}}}]$, where the convolution window size is from 1 to 4.
There are $\frac{1}{4}d_f$ convolution filters for each window size, and the stride length is 1. 
The zero-padding is used to let the input sequence and the output sequence of the convolution operation have the same length.
Thus, a history utterance $u_{m,k}$ has four corresponding matrices $\bm {U_{m,k}}^1, \bm{{U_{m,k}}^2}, \bm{{U_{m,k}}^3}, \bm{{U_{m,k}}^4}$ with a same dimension $\mathbb{R}^{n_{u_{m,k}}\times \frac{1}{4} d_f}$.
We perform a concatenation operation on the four matrices as the final representations of wording behavior, written by:
\begin{equation}
\bm {U_{m,k}}^c = ({\bm{U_{m,k}}^1}\oplus \bm{{U_{m,k}}^2}\oplus \bm{{U_{m,k}}^3}\oplus \bm{{U_{m,k}}^4})
\end{equation}
where ${\bm {U_{m,k}}^c} \in \mathbb{R}^{n_{u_{m,k}}\times d_f}$.
Accordingly, the wording behavior representation of a response is ${\bm {R_m}^c} \in \mathbb{R}^{n_r\times d_f}$ .
We further calculate the interaction matrix of ${\bm {U_{m,k}}^c}$ and ${\bm {R_m}^c}$
for capturing matching structure and patterns on wording behavior level.
Similar to the calculation of the interaction matrices elaborated in the last subsection, the $(p,q)$-th element of the $\bm{M_{m,k}}$ is calculated by the dot production of the $p$-th element of ${\bm {R_m}^c}$ and the $q$-th element of ${\bm {{U_{m,k}}^c}}$. In practice, we use matrix multiplications to calculate $\bm{M_{m,k}}$ as follows:
\begin{equation}
\bm{M_{m,k}}= {\bm {{R_m}^c}} \cdot {\bm {{U_{m,k}}^c}}^T
\end{equation}

\subsection{Aggregation and Fusion}
To aggregate matching degree information between a context utterance and a response, we alternatively stack two layers of 2-D convolution and max-pooling operation on the interaction matrices $\bm{{M_j}^{'}}, \bm{{M_j}^{1'}}, \bm{{M_j}^{2'}}, \bm{{M_j}^{3'}}, \bm{{M_j}^{a'}}$, where each interaction matrix is treated as an input channel, and the activation function is ReLU.
After this operation, a concatenation operation and an MLP with one hidden layer are used to flatten the output of the stacked CNN and generate a low-dimension vector for each context utterance response pair, denoted as $\bm{v_j}$.
As to the matching information aggregation between a history utterance and a response, we perform the same 2-D CNN with two layers on the interaction matrix $\bm{M_{m,k}}$.
After the concatenation and flatten layer, we obtain a vector $\bm{v_{m,k}}$
as the aggregation of $\bm{M_{m,k}}$. 
The dimensions of $\bm{v_j}$ and $\bm{v_{m,k}}$ are both $d_h$.
For multi-turn context-response matching, PHMN computes the aggregated matching vector between each utterance in context \begin{math}
c=(u_1, u_2, \dots, u_j, \dots, u_{n_c})
\end{math} and the corresponding response candidate $r$, resulting in a sequence of matching vectors $\bm{v_1}, \bm{v_2}, \dots, \bm{v_j}, \dots, \bm{v_{n_c}}$.
In matching between dialogue history and response, PHMN outputs a bag of matching vectors $\bm{v_{m,1}}, \bm{v_{m,2}}, \dots, \bm{v_{m,k}}, \dots, \bm{v_{m,n_m}}$ between each utterance in history 
\begin{math}
m=(u_{m,1}, u_{m,2}, \dots, u_{m,k}, \dots, u_{m,n_m})
\end{math} and the response candidate $r$.
Noticing that utterances in a context have a temporal relationship, we thus leverage an RNN with GRU cell to process the aggregated matching vectors $\bm{v_1}, \bm{v_2}, \dots, \bm{v_j}, \dots, \bm{v_{n_c}}$ and the use last state of RNN as the aggregated matching degree, namely $\bm{m^{rnn}}\in \mathbb{R}^{d_h \times 1}$.
On the other hands, utterances in dialogue history are parallel, and thus we use an attention mechanism \cite{bahdanau2014neural} to fuse the matching vectors $\bm{v_{m,1}}, \bm{v_{m,2}}, \dots, \bm{v_{m,k}}, \dots, \bm{v_{m,n_m}}$, i.e., computing the weighted sum as the aggregated matching degree, denoted as $\bm{m^{att}}\in \mathbb{R}^{d_h \times 1}$.
To facilitate the combination of context-response matching information and history-response matching degree, we leverage a dynamic gate mechanism \cite{tu2018learning}, which is formulated as follows:
\begin{equation}
\bm{\lambda}=\sigma({{\bm {U^{{rnn}}}}\bm{m^{rnn}}+{\bm {V^{{att}}}}\bm{m^{att}}})
\end{equation}
where $\bm{m^{rnn}}$ is the fused context-response matching degree and $\bm{m^{att}}$ corresponds to history-response matching, $\sigma$ represents the $sigmoid$ activation function.
The final combination of $\bm{m^{rnn}}$ and $\bm{m^{att}}$ is computed by
\begin{equation}
\bm{m^t}=(1-\bm{\lambda})\otimes \bm{m^{att}}+ \bm{\lambda}\otimes \bm{m^{rnn}}
\end{equation}
where $\otimes$ denotes element-wise multiplication.
$\bm{m^t}$ is then processed by a fully connected layer followed by a softmax function to obtain a binary output.

\subsection{Training}
In learning the matching functions $f(\cdot)$, the objective is to minimize the cross-entropy with dataset $\mathcal{D}$, which can be formulated as:
\begin{equation}
\mathcal{L}=-\sum_{i=1}^Ny_ilog(f(c_i,m_i,r_i))+ (1-y_i)log(1-f(c_i,m_i,r_i))
\end{equation}

We also construct two auxiliary loss functions to enhance the training process.
The first loss function refers to learning the binary classification outputs only based on context-response matching information (the upper section of Figure 1), written by:
\begin{equation}
\mathcal{L}_1=-\sum_{i=1}^Ny_ilog(g_1(c_i,r_i))+ (1-y_i)log(1-g_1(c_i,r_i))
\end{equation}
while another loss function corresponds to outputting the binary results based on history-response matching information (the bottom part in Figure 1), formulated as:
\begin{equation}
\mathcal{L}_2=-\sum_{i=1}^Ny_ilog(g_2(m_i,r_i))+ (1-y_i)log(1-g_2(m_i,r_i))
\end{equation}
where $g_1(\cdot)$ and $g_2(\cdot)$ refer to the matching function of context-response and history-response respectively.

%% file: figure-1.tex
\begin{figure*}
\centering
\includegraphics[width=1\textwidth]{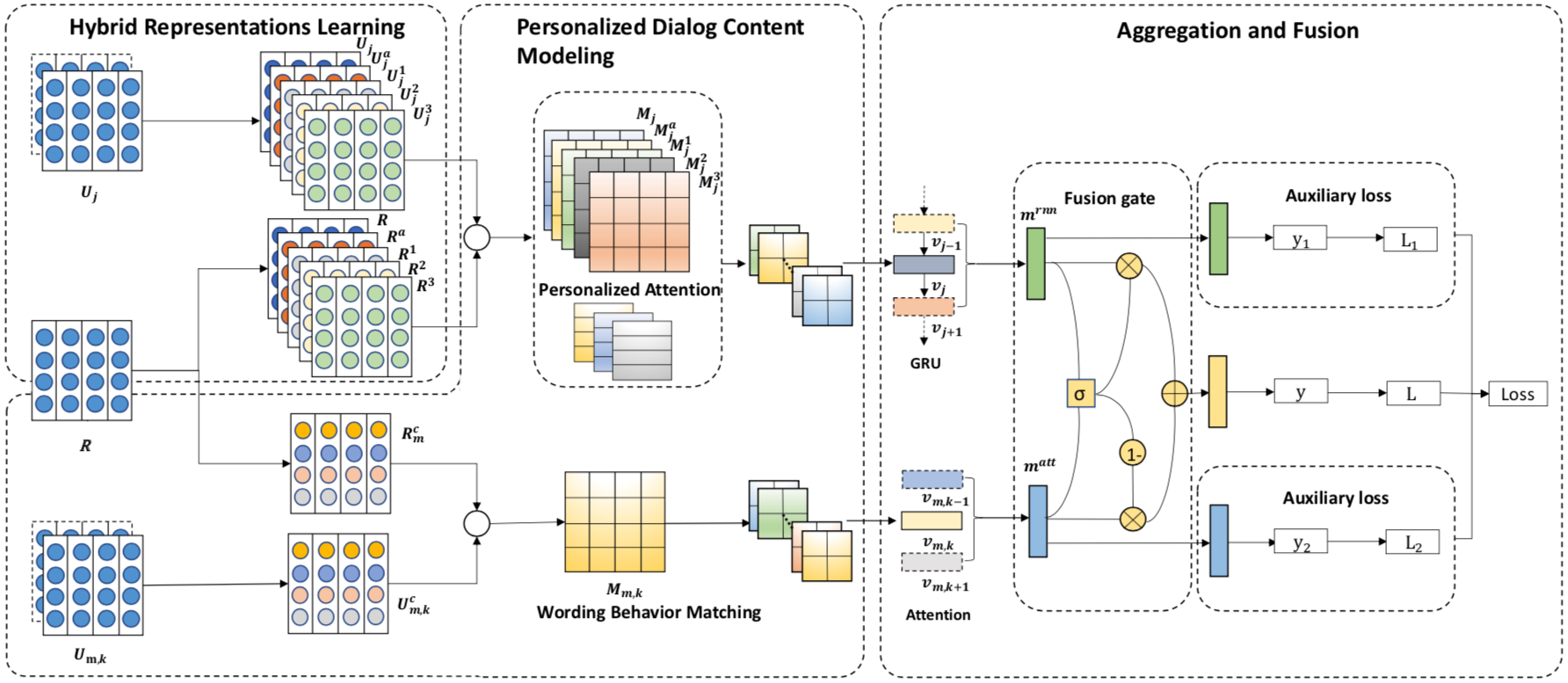}
\caption{\label{fig:framework}The detailed architecture of our PHMN model, which includes three parts, i.e., hybrid representation learning module, personalized dialogue content modeling, aggregation and fusion. }
\end{figure*}

%% file: figure-2.tex
\begin{figure}
\centering
\includegraphics[width=0.75\columnwidth]{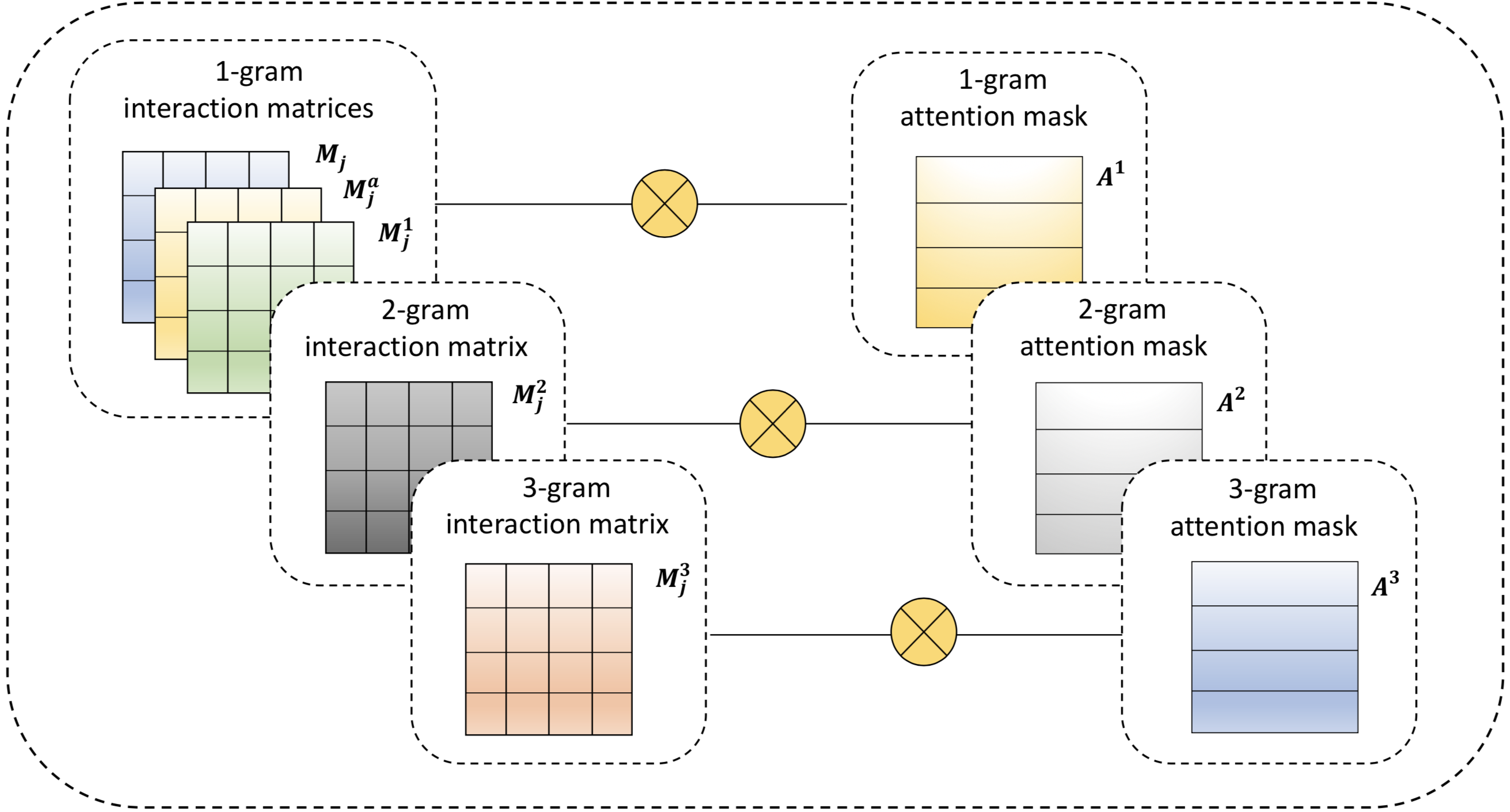}
\caption{\label{fig:mask}Details of the personalized attention over the hybrid representations between context and response.}
\end{figure}

%% file: 4-experiments.tex
\section{Experiments}

\input{table-2}

\subsection{Datasets}
To evaluate the effectiveness of our proposed model, we conduct experiments on two large open-datasets with user-id information, i.e., P-Ubuntu dialogue corpus in English, P-Weibo dataset in Chinese.
In detail, the P-Ubuntu dialogue corpus contains multi-turn technical support conversations with corresponded open user ids, which is collected from the Ubuntu forum \footnote{\url{https://ubuntuforums.org/}}.
We utilized Ubuntu v1.0 \cite{lowe2015ubuntu} as the raw dataset and followed the previous pre-processing strategy to replace numbers, paths, and URLs with placeholders \cite{xu2016incorporating}.
The P-Weibo corpus is crawled from an open Chinese online chatting forum \footnote{\url{https://www.weibo.com}} which contains massive multi-turn conversation sessions and user identification information. 

However, the traditional pre-processed version only contains context-response pairs, neglecting the user's ids and their dialogue history utilized in our proposed personalized ranking-based chatbots.
To mitigate this issue, we further process the raw dataset into a personalized version as follows.
We firstly filter out users who spoke less than 30 utterances in P-Ubuntu and 10 utterances in P-Weibo. The remaining users are considered as valid users, and we collect their utterances from the corresponding corpora as their dialogue history. The user's dialogue histories are truncated to the max length of 100 for P-Ubuntu and 50 for P-Weibo.
We then collect dialogue sessions of which the two speakers are both valid users from the raw corpora.
Next, we create dialogue cases from dialogue sessions by splitting them into several fragments each of which is comprised of several consecutive dialogue utterances. The last utterance in the fragment is considered as the gold response, and the remaining utterances are as the context.
To achieve this, we use a sliding window to split out dialogue cases from sessions. We set the maximum context turn to 10 for both corpora and the minimum context turn to 5 for P-Ubuntu and 3 for P-Weibo given their statistics. 
Furthermore, we pair each dialogue case with its users' information to facilitate the incorporation of personalized response selection, which contains the speaker's id, the speaker's dialogue history, the responder's id, the responder's dialogue history.
Note that for each pre-processed case, we make sure that the provided speaker's or the responder's dialogue history has no overlap with the dialogue session that the current dialogue case comes from to avoid information leakage.
Finally, after the aforementioned pre-processing steps, we get 600000 such six-point groups (context, response, speaker's id, speaker's dialogue history, responder's id, responder's dialogue history) as positive cases for both corpora. We randomly split them into 500000/50000/50000 for training/validation/testing.
For training, we randomly sample a negative response from other responses of the full dataset, so the proportion of the positive sample and the negative sample is 1:1. While for validation and testing, the number of randomly selected negative responses from the full dataset is 9 and the proportion is 1:9.
More statistical details of the two corpora are given in Table \ref{tab:data}.

\subsection{Baselines}
In our experiments, we compare our model with the following related and strong baselines.
Note that since we utilize two newly created datasets P-Ubuntu and P-Weibo, we run all these models by ourselves.

\textbf{TF-IDF}~\cite{lowe2015ubuntu}, a simple but effective matching method, computes the TF-IDF scores of each word in both context utterances and response. Both context utterances and responses are represented by their corresponding weighted addition of word embeddings. The matching score between context and response is then calculated by cosine similarity.

\textbf{LSTM}~\cite{lowe2015ubuntu} concatenates all utterances in the context into a long sentence and employs a shared LSTM network to convert both the context and the response into vector representations. Their matching degree is then calculated through a bi-linear function with sigmoid activation.

\textbf{Multi-View}~\cite{zhou2016multi} performs context-response matching calculation from multi-views, i.e., integrating information from both word sequence view and utterance sequence view to model two different levels of dependency.

\textbf{SMN}~\cite{wu2017sequential} refers to the sequential matching network. This framework separately processes each utterance in a given context to learn a matching vector between each utterance and the response with the CNN network. Then, the learned matching vectors are aggregated by RNN to calculate the final matching score between the context and the response candidate.

\textbf{DAM}~\cite{zhou2018multi}, the deep attention matching network, is a strong baseline for multi-turn response retrieval. This model builds a similar matching calculation pipeline upon the SMN, while the dependency between utterances in context and response candidates are captured by stacked self-attention and cross-attention mechanisms.

\textbf{MRFN}~\cite{tao2019multi} represents the multi-representation fusion network. The model performs context-response matching based on multiple types of sentence representations and fuses matching information from different channels effectively.

\textbf{IOI}~\cite{tao2019one} refers to the interaction-over-interaction network. This model performs deep-level matching by stacking multiple interaction blocks, i.e., extracting and aggregating the matching information within an utterance-response pair in an iterative fashion.

\textbf{MSN}~\cite{yuan2019multi} refers to the multi-hop selector network. This model firstly adopts a multi-hop selector to select the relevant utterances as context to avoid the side effect of using too many context utterances. Then, the model matches the candidate response with the filtered context to get a matching score.

\textbf{BERT$_{ft}$}~\cite{devlin2018bert} refers to the fine-tuned BERT-base model. This model is initialized with BERT-base-uncased and BERT-base-Chinese for P-Ubuntu and P-Weibo respectively. It takes the concatenation of the context and the candidate response as the input and utilizes stacked self-attention layers to extract fine-grained representations. The matching score is calculated with an MLP built upon the top layer.

\subsection{Experimental Settings}
We introduce the experimental settings in this subsection. Unless otherwise stated, the pre-processing methods and the hyperparameters are the same for both corpora.
We construct a shared vocabulary for context utterances, history utterances and responses, which contains the 30000 most frequent words on the training sets.
We then run Word2Vec\footnote{\url{https://code.google.com/archive/p/word2vec/}} on the training sets of the two corpora with the dimension of the word embedding as 200. 
Following previous work, we limit the length of context to 10 turns and truncate all context utterances to the max length of 50. 
As to user dialogue histories, we provide up to 100 user utterances for P-Ubuntu dataset and 50 sentences for P-Weibo dataset respectively.
If the number of turns in a context and the number of utterances in a user dialogue history have not reached the given upper limit, we append blank sentences whose words are all padding tokens.
In the hybrid representations learning of context-response matching module, we set the number of filters $d_w$ as 200 for \{1, 2, 3\}-gram CNN, and the number of heads as 8 for multi-head self-attention.
In the personalized dialogue content modeling part, we choose 50 as the filter number for all \{1, 2, 3, 4\}-gram CNN.
In the aggregation stage, the window sizes of 2-D convolution and pooling are (3, 3) for both context-response and history-response interactions. 
The dimension of the hidden state of the turn-level aggregation GRU is 200.
For training, we set the mini-batch size to 60 and adopt the Adam optimizer \cite{kingma2014adam} with the initial learning rate set to 3e-4. We exponentially decay the learning rate with the decay rate as 0.95 for every 2000 training steps. We utilize early stopping as a regularization strategy. 
The model which achieves the best performance on the validation set is used for testing.

For baseline models, we adopt their released codes if possible or implement ourselves and experiment on our proposed datasets. We ensure that all of our implemented baseline models achieve similar results as reported in the original papers in the standard Ubuntu v1.0 corpus.
These models utilize the same vocabulary and initial word embeddings as our model.
Specifically, for BERT$_{ft}$, we use BERT-base-uncased \footnote{\url{https://huggingface.co/bert-base-uncased}} for P-Ubuntu and BERT-base-Chinese \footnote{\url{https://huggingface.co/bert-base-chinese}} for P-Weibo respectively.
We first truncate the response to the max length of 50 and then iteratively insert the context utterances in reverse order before the response until we exhaust the context, or the total sequence exceeds the max sequence length of BERT (i.e., 512). We fine-tune the model using Adam optimizer \cite{kingma2014adam} with the learning rate of 3e-5 and the batch size of 32.

\subsection{Evaluation Metrics}
Given the candidate responses for each context of the test set, we evaluate the performance of different models with $R_{n}@ks$, which denotes whether top-$k$ retrieved responses from $n$ candidates contain the positive response. 
Besides, we also provide the top-$k$ ranking list for each test context to calculate the mean reciprocal rank (MRR) score, which is computed as follows:
\begin{equation}
    \text{MRR} = \frac{1}{\arrowvert \mathcal{T} \arrowvert} \sum_{ \langle c,m\rangle \in \mathcal{T}} \frac{1}{rank(\langle c,m\rangle)}
\end{equation}
where $\mathcal{T}$ indicates the context set for testing, $rank(\langle c,m\rangle)$ is the position of the true response regarding to the input $\langle c,m\rangle$ in the candidate ranking list. 

%% file: table-2.tex
\begin{table}[t]
\begin{center}
\caption{\label{tab:data} The statistical results of two large open datasets used in the experiments, i.e., P-Ubuntu and P-Weibo; C, R, Sess refer to context, response, and dialogue session, respectively; \# C-R pairs and Avg \#utts per user represent the total number of context-response matching pairs and the average number of utterances in the dialogue history of a user.}
\begin{tabular}{l|c c c |c c c }
\hline
\bf Corpus  & \multicolumn{3}{c|}{\bf P-Ubuntu} & \multicolumn{3}{c}{\bf P-Weibo}\\
\hline
Subsets & \bf Train & \bf Valid &\bf Test & \bf Train & \bf Valid & \bf Test \\
\hline
\# C-R Pairs &  1000k &  500k & 500k &  1000k &  500k &  500k \\
\hline
Avg \# turns per Sess &  8.6 &  8.6 & 8.6 &  4.4 &  4.4 & \ 4.4 \\
\hline
Avg \# words per C &  99.8 &  100.1 & 99.4 &  37.3 &  37.2 &  37.3 \\
\hline
Avg \# words per R &  12.1 &  12.0 & 12.0 &  7.8 &  7.9 &  7.8 \\
\hline
Avg \# utts per user &  93.9 &  93.9 & 93.9 &  23.0 &  23.0 &  22.9 \\
\hline
Avg \# words per utt &  12.0 &  11.9 & 11.9 &  7.9 &  7.9 &  7.9 \\
\hline
\end{tabular}
\end{center}
\end{table}

%% file: 5-results.tex
\section{Results}
\input{table-3}
Table \ref{tab:overall} reports the results of baselines and our proposed methods on P-Ubuntu and P-Weibo dataset. 
Table \ref{tab:ablation} supplements the evaluation results of model ablation on two datasets.
We analyze these results from the following aspects.

\subsection{Main Performance}
Overall, our proposed PHMN model significantly outperforms all other models in all metrics and achieves the new state-of-the-art results on  P-Ubuntu dialogue corpus and P-Weibo dataset.
Especially for $R_{10}@1$, PHMN achieves significant improvement over the most strong model without using BERT and its variations, i.e., MSN, on both datasets (i.e., (78.2 v.s. 70.9) on P-Ubuntu Corpus and (74.5 v.s. 70.3) on P-Weibo dataset). Surprisingly, when compared with BERT$_{ft}$ baseline, our proposed PHMN (without the support of BERT and its variations) still obtain significantly better results on P-Ubuntu Corpus (78.2 v.s. 75.7) and   P-Weibo dataset  (74.5 v.s. 74.0).
For baseline models, TF-IDF, LSTM, and Multi-view only achieve fundamental performances on each dataset and metric.
Benefiting from the deep neural model in matching feature extraction and sequential modeling strategy, SMN performs much better than the previous three baseline models on both datasets.
With the enhancement of powerful attention mechanism and deep stacked layers, DAM not surprisingly yields substantial improvements over SMN, which confirms that the attention mechanism is powerful for learning dependency representations of conversations.
Through fusing multiple types of sentence representation, MRFN
yields substantial improvement over DAM on the both P-Ubuntu corpus and P-Weibo dataset.
Furthermore, IOI and MSN perform slightly better than MRFN, these models are the strongest baselines to date without BERT and its variations.
BERT$_{ft}$ improves the scores of different metrics over other baselines by a large margin, but with the cost of model complexity and time efficiency, where the details are shown in Table \ref{tab:model_complexity}.

Our proposed HMN achieves comparable results with DAM by taking advantage of attention-based representations and interaction-based matching.
Considering that HMN contains only three convolution layers while DAM stacks multiple attention layers, the hybrid representations are thus time-efficient and effective.
Moreover, we notice that the simplified versions of PHMN, i.e., HMN$_{W}$ and HMN$_{Att}$, outperform MRFN and IOI on both corpora by a large margin. 

\input{table-4}

\subsection{The Effect of Wording Behavior}
As mentioned previously, the wording behavior is introduced for modeling long-term personal information other than the current dialogue context so as to enhance the performance of response candidate selection. We conduct the following two groups experiments, i.e., HMN$_W$ v.s. HMN, PHMN v.s. HMN$_{Att}$ as ablation studies to investigate how the wording behavior extracted from user dialogue history affects the response selection results.
HMN is the simplified version of PHMN without containing wording behavior modeling and personalized attention module.
HMN$_W$ boosts HMN with wording behavior modeling as extra hints for selecting response candidates, whereas HMN$_{Att}$ enhances HMN with the personalized attention module to extract important information from context utterances.
PMN only takes dialogue history utterances and response as input to extract wording behavior matching patterns and degrees.

{\textit{Using wording behavior information alone yields a relatively inferior matching performance.}}
As demonstrated in Table \ref{tab:overall}, PMN achieves a basic performance in terms of various matching accuracy.
PMN yields a better result than TF-IDF and insignificantly worse performance than the recently proposed models (i.e., LSTM and Multi-View) on the P-Ubuntu dialogue corpus.
It is also obtained that PMN is marginally better than LSTM and Multi-View on the P-Weibo dataset.
However, there is a significant gap between PMN and state-of-the-art models.
These results support the intuition that context utterances contain most of the patterns and information regarding selecting a proper response while the wording behavior models general and long-term matching information. 

{\textit{Wording behavior significantly enhances context-response matching network by introducing supplementary matching information.}}
Note again that wording behavior in dialogue history serves as the long-term information and can be utilized to supplement the short-term information in context utterances.
Not surprisingly, HMN$_W$ achieves a significant improvement over the HMN model, and even achieves a significant improvement over the MRFN, IOI, and MSN models, which are very strong among these models without incorporating BERT and its variations.
With the enhancement of wording behavior information, our proposed PHMN yields an observable improvement over HMN$_{Att}$ and obtains the new state-of-the-art on two large datasets.
These results confirm that wording behavior matching between user-specific dialogue history and response candidate is effective for multi-turn response selection.

\subsection{The Influence of Personalized Attention}
As previously stated, introducing the personalized attention module is expected to bring a positive effect on extracting important information in context-response matching. 
We investigate the influence of personalized attention with two groups of comparison, i.e., HMN$_{Att}$ v.s. HMN, PHMN v.s. HMN$_{W}$. 
Following observations are made in this investigation, which confirms that personalized attention is an effective add-on to the existing context-response matching model.

\input{table-5}

{\textit{Personalized attention module effectively improves the accuracy of context-response matching through extracting important information in context-response matching.}}
When personalized attention is introduced, HMN$_{Att}$ and PHMN model are capable of extracting meaningful matching information from the interaction matrices of context utterances and response while allocating less weight to unrelated matching signals.
As illustrated by the evaluation results in Table \ref{tab:overall}, personalized attention can substantially improve the performance of HMN and HMN$_{W}$.

{\textit{Performance improvement achieved by using personalized attention is less than by modeling wording behavior in dialogue history.}}
Recall that we propose to employ user-specific dialogue history content from two different perspectives, i.e., wording behavior and personalized attention.
It is natural to compare the effectiveness of personalized attention and wording behavior.
As illustrated in Table \ref{tab:overall}, personalized attention results in a substantial improvement over base models on two corpora while wording behavior achieves a significant improvement on two corpora, which indicates that wording behavior modeling is more important than personalized attention.

\subsection{The effect of Fusion Gate and Auxiliary Loss}
Table \ref{tab:ablation} summarizes the evaluation results of eight model variations so as to investigate the effect of the auxiliary loss and the gate mechanism.
We have the observation that, for both PHMN and HMN$_W$, the auxiliary loss is helpful for training on two corpora.
For PHMN and HMN$_W$, when adding the gate mechanism, it is not surprised that observable performance improvement is achieved.
We believe the improvement is partly because wording behavior information in dialogue history is not at the same level with hybrid representations while the gate mechanism can effectively balance the distinctions between different levels of representations.

\subsection{The effect of Dialogue History Size on Model Performance}
In our proposed PHMN model, the dialogue histories are used to calculate the personalized attention mask and perform wording behavior matching with the candidate response. 
On the one hand, we often don't have enough history utterances from the same user in some scenarios. On the other hand, there is a trade-off between speed and model performance.
Therefore, we study how the size of dialogue history influences the model performance in this subsection and leave the comparison of inference speed together with baselines to the next subsection.

As illustrated in Table \ref{tab:history_length}, we set the number of utterances in the dialogue history of the P-Weibo dataset to \{10, 20, 30, 40, 50\} and set the number of utterances in the dialogue history of the P-Ubuntu dataset to \{10, 30, 50, 70, 100\} for studying the influence of dialogue history size on model performance. 
It can be observed that even the available number of dialogue history is small (i.e., 10 and 30 utterances), all the three models can still yield a considerable improvement over the HMN baseline.
And with the increase of dialogue history size, all the models' performance continues to improve and is not saturated under the current limitation.
We can reasonably expect that with more dialogue histories available, PHMN will bring us more surprises.


\begin{table}[ht]
\begin{center}
\caption{\label{tab:model_complexity} Comparison of model size and inference speed.}
\resizebox{\columnwidth}{!}{
\begin{tabular}{l|cccccccc }
\hline
 &  LSTM& Multi-View & SMN & DAM & MRFN & IOI &  MSN & BERT$_{ft}$ \\
\hline
\# Params(M) &  6.3 & 6.7 & 6.4 & 8.1 &  9.6 &  15.3 & 7.6  &110 \\
\hline
\ Latency(ms) &  0.882 & 0.894 & 0.364 & 2.526 & 1.818 & 4.416 & 0.822  &17.2 \\
\hline 
 & HMN &  HMN$_{Att}$ &  HMN$_{W,100}$ & PHMN$_{10}$ &  PHMN$_{30}$ & PHMN$_{50}$ & PHMN$_{70}$ &  PHMN$_{100}$   \\
\hline
\# Params(M) &  6.7 &6.7 & 7.6 & 7.6 & 7.6 &  7.6 &  7.6 & 7.6   \\
\hline
\ Latency(ms) &  0.642 &0.648 &1.824 & 0.796 & 1.030 & 1.230 & 1.452 & 1.834  \\
\hline
\end{tabular}
}
\end{center}
\end{table}


\subsection{Comparison of Model Complexity}
Moreover, we also study the time and memory cost of our models by comparing them with baselines in terms of the number of model parameters and inference latency, which is measured as the average time cost of evaluating a single case on the same RTX 2080Ti GPU.

For our proposed models, we study HMN, HMN$_{Att}$, HMN$_{W}$, and PHMM. Recall that the model architectures are the same (i.e., HMN and HMN$_{Att}$, HMN$_{W}$ and PHMM) for some of our models, and the inference latency is relevant to the user dialogue histories for some of our models (i.e., HMN$_{W}$ and PHMM).
To be more specific, HMN$_{Att}$ shares the same model architecture with HMN, and the computation cost of the introduced personalized attention mechanism is agnostic to the number of user dialogue histories. While HMN$_{W}$ has the same model architecture as PHMN, both models' inference latency increases with the number of user dialogue histories. Thus, we also give the inference latency of PHMN models with different user dialogue history sizes (denoted as the subscript number).

The comparison results are illustrated in Table \ref{tab:model_complexity}.
Comparing our proposed models with baselines, we can easily conclude that PHMN is both time- and memory-efficient while performing remarkably well.
In terms of parameter size, it can be observed that our proposed PHMN model has similar parameters as the start-of-the-art non-BERT baseline MSN and is smaller than MRFN and IOI, not to mention BERT$_{ft}$ (which is 14.5 times larger than PHMN). This indicates that the significant performance improvement of our proposed model comes from the introduced strategies in this paper rather than a larger model size. 
When it comes to inference latency, we can find that PHMN is similar to MRFN and is 2.4 times faster than IOI. BERT$_{ft}$ again significantly drags on the group (which is 9.4 times slower than PHMN).

We then compare our proposed models. Comparing HMN with HMN$_{Att}$ or comparing HMN$_{W,100}$ with PHMN$_{100}$, we can find that the personalized attention mechanism is quite time-efficient as it almost adds no additional time cost.
As for the influence of dialogue histories on the inference speed, it can be seen that the latency increases linearly with the number of used dialogue histories, which poses a trade-off between speed and performance that can be tuned to tailor the application scenarios.

\subsection{Case Study}
\input{table-6}
In addition to evaluating PHMN with quantitative results, we also conduct case studies to illustrate its superiority over baselines.
Table \ref{tab:case_wording_behavior}  and Table \ref{tab:case_personalized_attention}  illustrate two examples of context-response matching with the enhancement of user-specific dialogue history modeling. 
Further, the tables also give the predictions of our proposed PHMN and two strong baselines (i.e., MSN and BERT$_{ft}$), with which we can better understand the superiority of PHMN.

For the example in Table \ref{tab:case_wording_behavior}, it is clearly shown that wording behavior is helpful for retrieving the correct response, i.e., ``\textbf{\textit{I've told you...}}'' in dialogue history can serve as the supplementary information other than context-response matching. 
From the models' prediction scores we can observe that all the models provide a high matching score to the first negative case, which not only has a large word overlap with the context (i.e., ``\textbf{\textit{man apt-get}}'') but also seems to have a plausible tone to respond to the last context utterance ``\textbf{\textit{What IS THE COMMAND}}'' though it ignores the remaining context information. Both MSN and BERT$_{ft}$ rank this negative response as the most appropriate response against all the 10 candidate responses, including the ground truth response. And our proposed PHMN successfully ranks the ground-truth response on top thanks to the wording behavior model mechanism that effectively captures supplementary information.

Example in Table \ref{tab:case_personalized_attention} reveals the effectiveness of personalized attention mechanism in extracting accurate information from the interaction matrices of context utterances and response.
By allocating a large weight to the key clue word ``\textbf{\textit{xfce4}}'' in response, the matching accuracy is enhanced. 
Again, it can be seen from the models' prediction scores that although all the three models rank the ground-truth response on top, the prediction scores of the first negative candidate response given by MSN and BERT$_{ft}$  is not low.
Meanwhile, PHMN assigns a high matching score to the ground truth response and a relatively low matching score to the first negative candidate response.
The gap between the top-ranked score and the second-ranked score of PHMN is much larger than that of BERT$_{ft}$ (0.80 v.s. 0.36) and MSN (0.80 v.s. 0.18), which indicates that our proposed PHMN is much more confident to select the ground-truth response. This superiority is owed to the personalized attention mechanism that highlights the key clue word ``\textbf{\textit{xfce4}}''.

It is observed that there are also inferior context-response matching cases in experiments.
A notable example pattern is that the extracted literal wording behavior information might overwhelm other informative words and structured knowledge in the dialogue history.
One potential solution for addressing such an issue is to enhance PHMN with fine-grained personalized information modeling and structured knowledge extraction.
We also notice that there exist a few extraordinary bad cases where both wording behavior and personalized attention introduce noise signals for context-response matching.
We believe this is due to the limited size of the dialogue history.
These phenomena and analyses point out the direction of potential future work.
\input{table-7}

\subsection{Study of Speaker's Persona Information}
We also consider incorporating the speaker's persona information into our proposed personalized response selection model to find whether it can help the model learn better and make the conversation more engaging.
Specifically, we assign persona embeddings that capture high-level persona features(i.e., topics, talking preferences, and so on) for speakers whose occurrences in the processed datasets are larger than a lower threshold (named \textbf{User Occurrence Threshold}).
We fuse the speaker's persona embedding into the context-response matching process to provide a speaker-aware matching vector for better capturing the speaker's preference.
We borrow the iterative mutual gating mechanism from Mogrifier LSTM \cite{melis2019mogrifier}, whose effectiveness has been verified, to allow the context-response matching vector and the speaker's persona embedding vector to mutually refine the useful information they carried. We name PHMN enhanced with the speaker's embedding as PHMN$_e$. Under this motivation, there could be many influential factors that might be crucial to the performance, here we mainly study four factors: (1) the gate position, (2) the number of mutual gating iterations, (3) the dimension of the persona embedding, and (4) the number of users who have persona embedding (which is closely related to \textbf{User Occurrence Threshold}).

For (1), we can perform mutual gating between the context-response matching vector and the speaker's persona embedding vector before or after the turn-level aggregation GRU. If the gate is before the turn-level GRU, the speaker's persona embedding can provide utterance-level guidance for original matching vectors $v_j$. We abbreviate this gate position as \textbf{Before}. If we inject the speaker's persona embedding after the turn-level GRU, it can guide the aggregated matching vector $m^{RNN}$ from a global perspective. We abbreviate this gate position as \textbf{After}. 
For (2), the gating iterations are set to \{1, 2, 3\} to study whether deep mutual interactions can boost the performance.
For (3), the dimension of persona embedding is set to \{50, 100, 200\}.
And we use the \textbf{User Occurrence Threshold} mentioned just before as the indicator for (4). Concretely, we set the \textbf{User Occurrence Threshold} to 
\{3, 4, 5\}, which means we only provide the speakers whose occurrences in the processed datasets are larger than \{3, 4, 5\} with a specific user embedding, while leaving other speakers with a shared UNK embedding. Under these settings, the numbers of remaining users for Ubuntu and Weibo are \{33303, 26228, 21059\} and \{54683, 24950, 12162\} respectively.

We conduct extensive experiments on the two corpora to determine whether incorporating the speaker's persona information is helpful. The experiment results are shown in Table \ref{tab:speaker_ablation}.
Unfortunately, we don't observe an improvement when taking the speaker's persona information into account, given additional computation and memory cost.
Specifically, on the Ubuntu corpus, all attempts fail to obtain better performance, while on Weibo, some settings of PHMN$_e$ get comparable performance with PHMN. Another interesting observation is that the less interaction iteration is, the smaller the persona embedding dimension is, the larger the lower threshold of user occurrence is, the better performance (also the smaller performance drop on Ubuntu dataset) the model PHMN$_e$ can achieve.
The above observations indicate that incorporating the speaker's persona information brings little benefit but more computation and memory cost. Thus we don't involve the speaker's persona information in our model. 
Nevertheless, the speaker's preference may still be beneficial to response selection in some scenarios. We leave the study of the incorporation of the speaker's persona information as one of our future works.

\input{table-8}

%% file: table-3.tex
\begin{table*}[t]
\begin{center}
\caption{\label{tab:overall} Experiment results on P-Ubuntu and P-Weibo datasets, where numbers in bold means the best performance of each metric. HMN, PMN, HMN$_W$, HMN$_{Att}$ represent the simplified versions of PHMN for ablation study. We run three times of these models with different initialized parameters to calculate p-value. Numbers marked with * mean that the improvement is statistically significant compared with the baseline (t-test with p-value < 0.01).}
\renewcommand\arraystretch{1.15}
\resizebox{\columnwidth}{!}{
\begin{tabular}{l|c c c c c |c c c c c }
\hline
 & \multicolumn{5}{c|}{\bf P-Ubuntu Corpus} & \multicolumn{5}{c}{\bf P-Weibo Corpus}\\
\cline{2-11}
& $R_{2}@1$ &  $R_{10}@1$ &  $R_{10}@2$ &  $R_{10}@5$ & MRR & $R_{2}@1$ &  $R_{10}@1$ &  $R_{10}@2$ &  $R_{10}@5$ &  MRR \\
\hline
TF-IDF~\cite{lowe2015ubuntu} & 71.8 & 29.9 & 46.6 & 77.1 & 49.6 & 70.4 &28.2 & 44.6 & 75.5  & 48.1\\
LSTM~\cite{lowe2015ubuntu} & 89.2 & 61.0 & 77.3 & 94.3 & 74.8     &88.2 & 59.3 & 75.7 & 93.3 &73.4\\
Multi-View~\cite{zhou2016multi} & 90.0 & 62.4 & 79.0 & 95.1 & 75.9 & 88.9 & 60.4 & 76.7 & 93.9 & 74.3 \\
SMN~\cite{wu2017sequential}        & 91.2 & 66.0 & 81.2 & 95.7 & 78.3   & 90.0 & 64.4 & 79.5  & 94.7 & 77.0\\
DAM~\cite{zhou2018multi}        & 91.7 & 68.6 & 82.4 & 96.0 & 79.9   & 91.2 & 68.0 & 81.7  & 95.5 & 79.4\\
MRFN~\cite{tao2019multi}        & 92.3 & 70.5 & 83.8 & 96.3 & 81.2   & 91.7 & 69.3 & 82.7  & 96.0 & 80.4 \\
IOI~\cite{tao2019one}        & 92.5 & 70.7 & 83.9 & 96.4 & 81.3   & 92.0 & 70.1 & 83.1  & 96.0 & 80.8 \\
MSN~\cite{yuan2019multi}        & 92.6 & 70.9 & 84.0 & 96.5 & 81.5  
& 92.1 & 70.3 & 83.2  & 96.1 & 81.0  \\
BERT ~\cite{devlin2018bert}      & 94.0 & 75.7 & 87.2  & 97.2 & 84.7 & 93.6 & 74.0 & 86.3 & 97.0 & 83.6 \\
\hline
\bf PHMN & \textbf{95.2}* & \textbf{78.2}* & \textbf{89.7}* & \textbf{98.5}* & \textbf{86.7}*   & \textbf{94.0}* & \textbf{74.5}* & \textbf{87.0}*  & \textbf{97.4} & \textbf{84.2}* \\
\hline
PMN & 88.5 & 57.0 & 74.8 & 94.7 & 72.3   & 88.0 & 58.4 & 74.5  & 93.2 & 72.6\\
HMN & 91.7 & 68.7 & 82.5 & 96.0 & 80.0   & 91.0 & 67.7 & 81.4  & 95.3 & 79.1\\
HMN$_{W}$ & 94.6 & 76.1 & 88.3 & 98.1 & 85.3   & 93.3 & 73.2 & 85.8  & 97.0 & 83.2\\
HMN$_{Att}$ & 93.9 & 74.0 & 86.6 & 97.6 & 83.8   & 92.3 & 70.1 & 83.7  & 96.4 & 81.0\\
\hline
\end{tabular}}
\end{center}
\end{table*}

%% file: table-4.tex
\begin{table*}[t]
\begin{center}
\caption{\label{tab:ablation} Ablation study results for fusion gate and auxiliary loss on P-Ubuntu and P-Weibo corpora. The models with subscript $_{gate}$ use fusion gate, otherwise, they simply concatenate the two matching vectors and make a binary classification via a fully connected layer. The models with subscript $_{\mathcal{L+L}_1+\mathcal{L}_2}$ are additionally equipped with auxiliary loss. Our full model PHMN mentioned above adopts both fusion gate and auxiliary loss, for clarity, we denoted it as PHMN$_{\mathcal{L+L}_1+\mathcal{L}_2+gate}$ here.} 
\renewcommand\arraystretch{1.15}
\resizebox{\columnwidth}{!}{
\begin{tabular}{l|c c c c c |c c c c c }
\hline
 & \multicolumn{5}{c|}{\bf P-Ubuntu Corpus} & \multicolumn{5}{c}{\bf P-Weibo Corpus}\\ \cline{2-11}
& $R_{2}@1$ &  $R_{10}@1$ &  $R_{10}@2$ &  $R_{10}@5$ & MRR & $R_{2}@1$ &  $R_{10}@1$ &  $R_{10}@2$ &  $R_{10}@5$ &  MRR \\
\hline
PHMN$_{\mathcal{L}}$ &94.8 & 76.7 & 88.9 & 98.3 & 85.7    & 93.4 & 73.3 & 86.0  & 97.1 & 83.3\\
PHMN$_{\mathcal{L}+gate}$ & 94.9 & 77.2 & 89.1 & 98.4 & 86.0   & 93.6 & 73.8 & 86.3  & 97.2& 83.6 \\
PHMN$_{\mathcal{L+L}_1+\mathcal{L}_2}$  & 95.1 & 77.8 & 89.5 & 98.4 &86.4   & 93.6 & 74.1 & 86.5  & 97.3 & 83.8 \\
PHMN$_{\mathcal{L+L}_1+\mathcal{L}_2+gate}$ & \textbf{95.2} & \textbf{78.2} & \textbf{89.7} & \textbf{98.5} & \textbf{86.7}   & \textbf{94.0} & \textbf{74.5} & \textbf{87.0} & \textbf{97.4} & \textbf{84.2} \\
\hline
HMN$_{W+\mathcal{L}}$   & 94.1 & 75.0 & 87.8 & 97.9 & 84.6      & 93.0 & 72.2 & 85.3 & 96.8 & 82.5 \\
HMN$_{W+\mathcal{L}+gate}$ & 94.3 & 75.4 & 88.0 & 98.0 & 84.9         &93.1 & 72.6 &85.4 &96.9 & 82.7 \\
HMN$_{W+\mathcal{L+L}_1+\mathcal{L}_2}$   & 94.5 & 75.8 & 88.3 & 98.1 & 85.1         &93.2 & 72.8 & 85.7 & 97.0 & 82.9 \\
HMN$_{W+\mathcal{L+L}_1+\mathcal{L}_2+gate}$ & 94.6 & 76.1 & 88.3 & 98.1 & 85.3   & 93.3 & 73.2 & 85.8 & 97.0 & 83.2 \\
\hline
\end{tabular}}
\end{center}
\end{table*}

%% file: table-5.tex
\begin{table*}[t]
\begin{center}
\caption{\label{tab:history_length}Model performance of different numbers of utterances in the dialogue history of users. }
\renewcommand\arraystretch{1.15}
\resizebox{\columnwidth}{!}{
\begin{tabular}{l|c|ccccc|ccccc}
\hline
\multirow{2}{*}{} & \multirow{2}{*}{\bf \#Utts (M)} & \multicolumn{5}{c|}{\bf P-Ubuntu Corpus}   & \multicolumn{5}{c}{\bf P-Weibo Corpus}    \\ 
\cline{3-12} &  & $R_{2}@1$ & $R_{10}@1$ & $R_{10}@2$ & $R_{10}@5$ & MRR  & $R_{2}@1$ & $R_{10}@1$ & $R_{10}@2$ & $R_{10}@5$ & MRR  \\ 
\hline
\multirow{6}{*}{HMN$_{Att}$} 
& 0   & 91.7 & 68.7  & 82.5 & 96.0  & 80.0   & 91.0  & 67.7 & 81.4 & 95.3 & 79.1 \\ 
& 10  &92.6  &70.1  &83.6  &96.7  &81.0     &91.6 & 68.8 & 82.6 &96.0  &80.0 \\
& 30  &93.1  &71.8  &85.1  &97.0  &82.3     &92.0 & 69.7 & 83.4 & 96.3 &80.7 \\
& 50  &93.4  &72.7  &85.8  &97.2  &82.9     & 92.3 & 70.1 & 83.7  & 96.4 & 81.0 \\
& 70  &93.7  &73.4  &86.2  &97.4  &83.4     &- &- &- &- &- \\
& 100  & 93.9 & 74.0  & 86.6 & 97.6  & 83.8   &- &- &- &- &- \\ 
\hline
\multirow{6}{*}{HMN$_{W}$} 
& 0   & 91.7 & 68.7  & 82.5 & 96.0  & 80.0   & 91.0  & 67.7 & 81.4 & 95.3 & 79.1 \\ 
& 10  &93.0  &71.7  &85.3  &97.1  &82.2     & 92.6 & 71.5 & 84.6 & 96.5 & 81.8 \\
& 30  &93.6  &73.4  &86.2  &97.3  &83.3     & 93.1  &72.7 & 85.7 & 96.9 & 82.9 \\
& 50  &94.0  &74.6  &87.3  &97.6  &84.3     & 93.3  & 73.2 & 85.8 & 97.0 & 83.2 \\
& 70  &94.5  &75.6  &88.1  &98.0  &85.0     &- &- &- &- &- \\
& 100  & 94.6 & 76.1  & 88.3 & 98.1  & 85.3   &- &- &- &- &- \\
\hline
\multirow{6}{*}{PHMN} 
& 0   & 91.7 & 68.7  & 82.5 & 96.0  & 80.0   & 91.0  & 67.7 & 81.4 & 95.3 & 79.1 \\ 
& 10  &93.8  &73.5  &86.5  &97.6  &83.5     &93.1 & 72.6 & 85.4 &97.0  &82.7 \\
& 30  &94.4  &75.5  &87.9  &98.1  &84.9     & 93.6 &  74.0 & 86.6 & 97.3 &83.7 \\
& 50  &94.8  &76.8  &88.8  &98.2  &85.8     & 94.0  & 74.5 & 87.0 & 97.4 & 84.2 \\
& 70  &95.1  &77.6  &89.3  &98.4  &86.3     &- &- &- &- &- \\
& 100  &95.2 &78.2 & 89.7 &98.5 &86.7     &- &- &- &- &- \\
\hline
\end{tabular}
}
\end{center}
\end{table*}

%% file: table-6.tex
\begin{table}[t]
\begin{center}
\caption{\label{tab:case_wording_behavior} A sampled case from P-Ubuntu corpus that shows the effectiveness of \textit{wording behavior} modeling.}
\begin{tabular}{c|c|ccc}
\hline
& \multicolumn{4}{l}{B: as \textcolor{blue}{\textbf{\textit{I've told you}}} 3 times now}\\ 
{\bf Dialogue } & \multicolumn{4}{l}{B:  \textcolor{blue}{\textbf{\textit{I've told you}}} 3 times, install "mysql-server"}\\ 
{\bf History }& \multicolumn{4}{l}{B: \textcolor{blue}{\textbf{\textit{I've told you}}} 2 times there are guides on https://help.ubuntu.com}\\ 
& \multicolumn{4}{l}{B:  \textcolor{blue}{\textbf{\textit{I've told you}}} what to do, and \textcolor{blue}{\textbf{\textit{I've told you}}} this is not an ubuntu support issue}\\ 
\hline
\multirow{6}*{\bf Context}
& \multicolumn{4}{l}{A: What's the command to order packages by size? I'd like the command line one } \\
& \multicolumn{4}{l}{and the GUI one} \\
& \multicolumn{4}{l}{B: look in the gui for sort functions. man dpkg and man apt-get} \\
& \multicolumn{4}{l}{A: please tell me what command to use} \\
& \multicolumn{4}{l}{B: man apt-get and man dpk will show you the commands} \\
& \multicolumn{4}{l}{A: What IS THE COMMAND} \\
\hline \hline

\multirow{2}*{\bf Label} &\multirow{2}*{\bf {Candidate Response}} & \multicolumn{3}{c}{\bf {Model Prediction}} \\ \cline{3-5}
\multirow{2}*{} &\multirow{2}*{} & \bf MSN & \bf {BERT$_{ft}$} & \bf PHMN \\
\hline

\multirow{2}*{\ding{52}} & \multicolumn{1}{l|}{\textcolor{red}{\textbf{\textit{I've told you}}} the command to find out the commands} & \multirow{2}*{0.32}&\multirow{2}*{0.48} &\multirow{2}*{0.79} \\
\multirow{2}{*}{}  & \multicolumn{1}{l|}{you want}  &\multirow{2}{*}{} &\multirow{2}{*}{} &\multirow{2}{*}{}\\
\hline
\multirow{10}*{\ding{56}}
& \multicolumn{1}{l|}{try 'man apt-get' and 'man aptitude'}  & 0.58 & 0.64 & 0.43 \\ \cline{2-5}
&\multicolumn{1}{l|}{I wanna eventually be a Unix Admin}  & 0.02 & \textless 0.01 & 0.01 \\ \cline{2-5}
&\multicolumn{1}{l|}{that should be good enough}  & 0.07 & 0.02 & 0.05 \\ \cline{2-5}
&\multicolumn{1}{l|}{so you'll have anonymous access and such.}  & 0.02 & 0.01 & \textless 0.01 \\ \cline{2-5}
&\multicolumn{1}{l|}{It's kind of like a menu... in expanded mode.}  & 0.08 & 0.04 & 0.03 \\ \cline{2-5}
&\multicolumn{1}{l|}{and it works good for the time being :)}  & \textless 0.01 & \textless 0.01 & \textless 0.01 \\ \cline{2-5}
&\multicolumn{1}{l|}{sounds like dissection's suggestion might do the trick.}  & 0.03 & \textless 0.01 & 0.01 \\ \cline{2-5}
&\multicolumn{1}{l|}{it may already be supported, have a websearch round}  & 0.05 & 0.02 & 0.03 \\ \cline{2-5}
&\multicolumn{1}{l|}{no, it is installed now.}  & 0.06 & 0.12 & 0.17 \\ \hline
\end{tabular}
\end{center}
\end{table}

%% file: table-7.tex
\begin{table}[ht]
\begin{center}
\caption{\label{tab:case_personalized_attention} A sampled case from P-Ubuntu corpus that shows the advantage of utilizing \textit{personalized attention}. 
}
\begin{tabular}{c|c|ccc}
\hline
& \multicolumn{4}{l}{B: i've read somewhere that \textcolor{blue}{\textbf{\textit{xfce4}}} is as fast as fluxbox}\\ 
{\bf Dialogue } & \multicolumn{4}{l}{B: i use \textcolor{blue}{\textbf{\textit{xfce4}}} , old laptop gnome runs terribly slow on it}\\ 
{\bf History }& \multicolumn{4}{l}{B: haven't tried kde on this laptop, but when i tried \textcolor{blue}{\textbf{\textit{xfce4}}} its like a new }\\ 
& \multicolumn{4}{l}{lease of life. \textcolor{blue}{\textbf{\textit{xfce4}}} is light, yet quite functional}\\ 
\hline
\multirow{7}*{\bf Context}
& \multicolumn{4}{l}{A: do anyone know how to add shortcuts to the menu? } \\
& \multicolumn{4}{l}{B: depends on your desktop environment} \\
& \multicolumn{4}{l}{A: sorry i new in ubuntu, what do you mean with desktop enviroment?} \\
& \multicolumn{4}{l}{B: KDE / GNOME / \textcolor{red}{\textbf{\textit{xfce4}}}/ fluxbox??} \\
& \multicolumn{4}{l}{A: its GNOME} \\
& \multicolumn{4}{l}{B: old laptop GNOME runs terribly slow on it} \\
& \multicolumn{4}{l}{A: umm yup.. what do you suggest then?} \\
\hline \hline

\multirow{2}*{\bf Label} &\multirow{2}*{\bf {Candidate Response}} & \multicolumn{3}{c}{\bf {Model Prediction}} \\ \cline{3-5}
\multirow{2}*{} &\multirow{2}*{} & \bf MSN & \bf {BERT$_{ft}$} & \bf PHMN \\
\hline

\multirow{2}*{\ding{52}} & \multicolumn{1}{l|}{Try \textcolor{red}{\textbf{\textit{xfce4}}} it's wonderfull, as light as icewm,} & \multirow{2}*{0.67}&\multirow{2}*{0.81} &\multirow{2}*{0.97} \\
\multirow{2}{*}{}  & \multicolumn{1}{l|}{and more comfortable to use}  &\multirow{2}{*}{} &\multirow{2}{*}{} &\multirow{2}{*}{}\\
\hline
\multirow{11}*{\ding{56}}

& \multicolumn{1}{l|}{have you tried GNOME's "network browser"?} & 0.49 & 0.45 & 0.17 \\ \cline{2-5}

& \multicolumn{1}{l|} {humm, well I dunno.  you check lsmod to see} & \multirow{2}*{0.01}&\multirow{2}*{\textless 0.01} &\multirow{2}*{\textless 0.01} \\
& \multicolumn{1}{l|}{what loaded and dmesg?}  &\multirow{2}{*}{} &\multirow{2}{*}{} &\multirow{2}{*}{}\\ \cline{2-5}

& \multicolumn{1}{l|}{no ,but I was hoping it is not necessary if you} & \multirow{2}*{0.04}&\multirow{2}*{0.02} &\multirow{2}*{0.05} \\

& \multicolumn{1}{l|}{use lagacy drivers} &\multirow{2}{*}{} &\multirow{2}{*}{} &\multirow{2}{*}{}\\ \cline{2-5}

& \multicolumn{1}{l|}{i quite love my intel graphics} & \textless 0.01 & \textless 0.01 & \textless 0.01 \\ \cline{2-5}

& \multicolumn{1}{l|}{but I can not ping anything} & \textless 0.01 & \textless 0.01 & \textless 0.01 \\ \cline{2-5}

& \multicolumn{1}{l|}{i made the steps and i have an error} & 0.02 & \textless 0.01 & 0.01 \\ \cline{2-5}

& \multicolumn{1}{l|}{its not exactly lubuntu but it works just the same} & 0.01 & 0.02 & 0.01 \\ \cline{2-5}

& \multicolumn{1}{l|}{im new to linux, i need more in depth infomation} & 0.04 & 0.13 & 0.09 \\ \cline{2-5}

& \multicolumn{1}{l|}{I'll check that out after I try to get DRI working.} & 0.11 & 0.07 & 0.03 \\ \hline
\end{tabular}
\end{center}
\end{table}

%% file: table-8.tex
\begin{table}[]
\caption{\label{tab:speaker_ablation} Study of incorporating speaker's persona information.}
\resizebox{\columnwidth}{!}{
\begin{tabular}{c|c|c|c|c|cc|cc}
\hline
\multirow{2}{*}{}       & \multirow{2}{*}{\textbf{\begin{tabular}[c]{@{}c@{}}Gate \\ Position\end{tabular}}} & \multirow{2}{*}{\textbf{\begin{tabular}[c]{@{}c@{}}Gating \\ Iterations\end{tabular}}} & \multirow{2}{*}{\textbf{\begin{tabular}[c]{@{}c@{}}Embedding \\ Dimension\end{tabular}}} & \multirow{2}{*}{\textbf{\begin{tabular}[c]{@{}c@{}}User Occurrence  \\ Threshold\end{tabular}}} & \multicolumn{2}{c|}{\textbf{Ubuntu}} & \multicolumn{2}{c}{\textbf{Weibo}} \\ 
\cline{6-9} 
 &  &   &   &  & R@1   & MRR  & R@1  & MRR \\ 
 \hline
PHMN  & -   & -  & -  & - & \textbf{78.2} & \textbf{86.7}  & \textbf{74.5} & \textbf{84.2}  \\ 
\hline
\multirow{11}{*}{PHMN$_e$} & Before  & \multirow{2}{*}{1}  & \multirow{2}{*}{50} & \multirow{2}{*}{5}  & 77.8 & 86.5  & 74.6 & 84.1 \\ 
\cline{2-2}
& After  &  &   & 77.7  & 86.4 & 74.4 & 83.9 \\ 
\cline{2-9} 
 & \multirow{3}{*}{Before} & 1   & \multirow{3}{*}{50}  & \multirow{3}{*}{5} & 77.8 & 86.5  & 74.6  & \multicolumn{1}{l}{84.1} \\ 
 \cline{3-3} 
 &  & 2  &  &  & 77.7  & 86.4  & 74.4 & 84.0 \\ 
 \cline{3-3}
& & 3&   &  & 77.3& 84.1  & 74.3  & 83.9   \\ 
\cline{2-9} 
& \multirow{3}{*}{Before}  & \multirow{3}{*}{1}  & 50 & \multirow{3}{*}{5}  & 77.8  & 86.5 & 74.6          & \multicolumn{1}{l}{84.1} \\ 
\cline{4-4}
& &  & 100  &  & 77.7 & 86.4  & 74.4 & 84.0  \\ 
\cline{4-4}
 &   &  & 200   &  & 77.3  & 86.2 & 74.3  & 84.0 \\ 
 \cline{2-9} 
& \multirow{3}{*}{Before}  & \multirow{3}{*}{1}  & \multirow{3}{*}{50} & 5  & 77.8 & \multicolumn{1}{l|}{86.5} & 74.6   & \multicolumn{1}{l}{84.1} \\ 
\cline{5-5}
&  &   &   & 4 & 77.6  & 86.3  & 74.5 & 84.1  \\ 
\cline{5-5}
&    &  &  & 3    & 77.4 & 86.3  & 74.4  & 84.0  \\ 
\hline
\end{tabular}
}
\end{table}

%% file: 6-related_work.tex
\section{Related Work}
In the past decades, human-machine conversation systems have been widely investigated and developed. Early studies mainly focus on building rules and templates for computers to yield a human-like response. Such a strategy has been evolved and successfully used in various domains, such as museum guiding \cite{ferguson1996trains}, restaurant booking \cite{lei2018sequicity}. 
Later on, with the explosive growth of data, the application of the open-domain conversation model is promising. 
However, conventional methods for domain-specific settings have obstacles to scale to open area. 
Given this, various data-driven approaches have been proposed for modeling open-domain conversation, including two main groups:  generation-based approaches \cite{serban2015building,chan2019modeling,qiu2019training,li2019insufficient,li2018overview}, retrieval-based methods\cite{yan2016learning}. 
Early work of the first group builds their systems upon statistical machine translation model \cite{ritter2011data}.
Recently, on top of the sequence to sequence architecture \cite{shangL2015neural,vinyals2015neural}, various extensions have been proposed to address the ``common response'' issue \cite{li2015diversity}; to leverage external knowledge \cite{mou2016sequence,serban2016multiresolution,xing2017topic}; to model the hierarchical structure of conversation contexts \cite{serban2016building,serban2017hierarchical,xing2018hierarchical}; to generate personalized responses \cite{li2016persona,zhou2018emotional}; and to pursue effective optimization strategies \cite{li2016deep,li2017adversarial}.

Early work for retrieval-based dialogue systems studies single-turn response selection \cite{hu2014convolutional,ji2014information,lu2013deep,wang2013dataset}.
Later on, various multi-turn response selection methods have been proposed, including the dual LSTM model, \cite{lowe2015ubuntu}, the multi-view matching method \cite{zhou2016multi}, the sequential matching network \cite{wu2017sequential}, and the deep attention matching network \cite{zhou2018multi}.
Recently, various effective methods have been proposed for investigating the fusion of multiple types of sentence representations \cite{tao2019multi}, the deep interaction in matching feature extraction \cite{tao2019one}, model ensemble \cite{zhang2019ensemblegan,yang2019hybrid,song2018ensemble}, external knowledge combination \cite{yang2018response}, the influence of stickers in multi-modal response selection \cite{gao2020learning}, and emotion control in context-response matching \cite{lisong2020emotion}.
With the rapid explosion of pre-trained language models, researchers also have made considerable efforts in combining pre-trained language models with response selection.
One typical method is to combine a pre-trained language model (BERT) with post-training method in the task of response selection\cite{whang2020effective}.
Gu et al. \cite{gu2020speaker} further investigate the problem of employing pre-trained language models for Speaker-Aware multi-turn response selection. 
Lu et al. \cite{lu2020improving} propose two strategies to improve pre-trained contextual language models for response retrieval in multi-turn conversation, namely speaker segmentation and dialogue augmentation.
A deep context modeling architecture (DCM) with BERT as the context encoder has also been proposed for multi-turn response selection \cite{li58deep}.   
To address the issue of ignoring the sequential nature of multi-turn dialogue systems in utilizing pre-trained language models, the utterance manipulation strategie (UMS) has been proposed \cite{whang2020response}.
Wang et al. \cite{wang2020response} propose an essential pre-training step to embed topic information into BERT with self-supervised learning in multi-party multi-turn response selection. 
More details of progresses and challenges in building intelligent open-domain dialogue systems can be found in recent surveys \cite{huang2020challenges,boussaha2019deep}.

In this work, we proposed a personalized hybrid matching network (PHMN) for multi-turn response selection.
We combine deep attention-based representations and interaction information as hybrid representations to achieve comprehensive modeling of multi-turn context utterances.
Besides, we introduce personalized dialogue history as additional information to enhance the accuracy of context-response matching.
Through extracting wording behavior and personalized attention weights from the dialogue history, our proposed PHMN achieves state-of-the-art performance on two datasets.